\documentclass{article}

\usepackage{arxiv}

\usepackage{authblk}
\usepackage{amsmath}
\usepackage{graphicx}
\usepackage{subcaption}
\usepackage{enumitem}

\usepackage{amssymb}
\usepackage{natbib}
\usepackage[table]{xcolor}
\usepackage{algorithm}
\usepackage{algcompatible}
\usepackage[pagebackref=true,breaklinks=true,colorlinks,bookmarks=false]{hyperref}
\usepackage[capitalise]{cleveref}
\definecolor{RED}{rgb}{1,0,0}
\definecolor{ORANGE}{rgb}{1,0.5,0}
\definecolor{BLUE}{rgb}{0,0,1}

\usepackage{amsmath,amsfonts,bm}

\def\eqref#1{equation~\ref{#1}}
\def\1{\bm{1}}

\DeclareMathAlphabet{\mathsfit}{\encodingdefault}{\sfdefault}{m}{sl}
\SetMathAlphabet{\mathsfit}{bold}{\encodingdefault}{\sfdefault}{bx}{n}

\newcommand{\acrotntorch}{\textsf{tntorch}}

\begin{document}

\title{\acrotntorch{}: Tensor Network Learning with PyTorch}
	
\author[1*]{Mikhail Usvyatsov}
\author[2*]{Rafael Ballester-Ripoll}
\author[1]{Konrad Schindler}
\affil[1]{ETH~Zurich, Switzerland, \{mikhailu, schindler\}@ethz.ch}
\affil[2]{IE University,~Madrid, Spain, rafael.ballester@ie.edu}
\affil[*]{Joint first authors}

\maketitle

\begin{abstract}%   <- trailing '%' for backward compatibility of .sty file
We present \acrotntorch{}, a tensor learning framework that supports multiple decompositions (including \textsc{Candecomp/Parafac}, Tucker, and Tensor Train) under a unified interface. With our library, the user can learn and handle low-rank tensors with automatic differentiation, seamless GPU support, and the convenience of PyTorch's API. %\acro{} allows for learning and handling tensors via PyTorch's automatic differentiation and seamless GPU support, and provides algorithms for decomposing data tensors into desired formats.
Besides decomposition algorithms, \acrotntorch{} implements differentiable tensor algebra, rank truncation, cross-approximation, batch processing, comprehensive tensor arithmetics,
%read and write operations using fancy indexing,
and more.
% Thanks to these features, the user can handle many decompositions as if they were pure uncompressed tensors.
\end{abstract}

\keywords{tensor decompositions \and pytorch \and low-rank methods \and multilinear algebra}

%\vspace{-.5em}
\section{Introduction}
%Many objects can be represented as $n$-dimensional regular grid of values.
%When $n$ is equal to 2 such object is called matrix, when $n > 2$ - tensor.
%Tensors memory complexity is exponential with respect to their dimensionality.
%Tensor decomposition methods allow for drop-in replacement of tensors with tensor networks often offering reduction either in memory requirement or in runtime of operations with tensors.

%Tensor network is a graph with data factors (smaller tensors, often called "cores") as nodes and matvec operations as edges. See \cref{fig:4dcp,fig:4dtttucker} for examples of 4d tensor decomposition into a tensor network.

%Tensor decompositions apply low-rank constraints to shared edges of the tensor network

In many machine learning and data analysis tasks one is faced with multi-dimensional data arrays. Tensors are a powerful tool to represent and handle such data, but often constitute a bottleneck in terms of storage and computation.
\emph{Tensor decompositions} expand a tensor into a set of separable terms. If the tensor has low rank (i.e., there are much fewer degrees of freedom than tensor elements), then such a decomposition can dramatically reduce the representation size~\citep{KB:09,CLOPZM:16,khrulkov2019tensorized}. 

Tensor decompositions play an increasingly important role in machine learning: they are used to factorize neural network weights~\citep{NPOV:15,gusak2019musco,wang2020kronecker,idelbayev2020low,kossaifi2020factorized}, to accelerate learning and inference~\citep{ma2019tensorized,hrinchuk2019tensorized,UMBRKS:21}, to impose priors that improve accuracy~\citep{kuznetsov2019prior,he2021hyperspectral} and robustness~\citep{kolbeinsson2021tensor}, etc. For more applications of tensors in machine learning see~\citep{panagakis2021tensor}. Importantly, the distinction between the different classical decomposition schemes has become blurred, as newer types of low-rank constraints and tensor indexing schemes have gained popularity~\citep{NPOV:15,khrulkov2019tensorized,CLOPZM:16,UMBRKS:21}. There exist multiple frameworks~\citep{NIKFO:20, KPAP:19} with emphasis on different aspects. We believe that more than ever, tensor software must be flexible regarding the underlying tensor formats, while exposing a natural and accessible interface to the user.

To this end, we introduce \acrotntorch{} (\url{github.com/rballester/tntorch}), an open-source Python package that abstracts the choice of format, while providing a wide range of tools for tensor learning, manipulation, and analysis. Compared to similar recent libraries T3F~\citep{NIKFO:20}, TensorLy~\citep{KPAP:19}, TedNet~\citep{PWX:21}, TensorD~\citep{HLYX:18}, TensorNetwork~\citep{RMGZFZHVL:19}, and tt-pytorch~\citep{khrulkov2019tensorized}, \acrotntorch{} emphasizes an easy-to-use, decomposition-independent interface inherited from PyTorch\footnotemark{}\footnotetext[1]{\acrotntorch~ can be installed conveniently via the PyPI package manager (command: \texttt{pip install tntorch})}.
%\acro~ is heavily dependent on NumPy~\citep{HMW:20}. The optional dependences CVXPY~\citep{DB:16}, SciPy~\citep{VGO:20}, and maxvolpy~\citep{maxvolpy} are required for least-squares problems, convolution, and cross-approximation, respectively.
The package's API is fully documented and features a score of tutorial Jupyter notebooks (\url{tntorch.readthedocs.io}) %\footnote{\label{readthedocs}\url{https://tntorch.readthedocs.io}.}
across a variety of use cases. See \cref{app:comparison} for a point-by-point comparison with the aforementioned software packages as of June 2022.

%\FIXME{See if we will make it work with JAX: \textcolor{blue}{we can, at least tensor construction is not that difficult to port}}

\section{Supported Tensor Formats}

\begin{figure}[h!]
	\centering
	\begin{subfigure}[b]{0.4\textwidth}
        \includegraphics[width=1\linewidth]{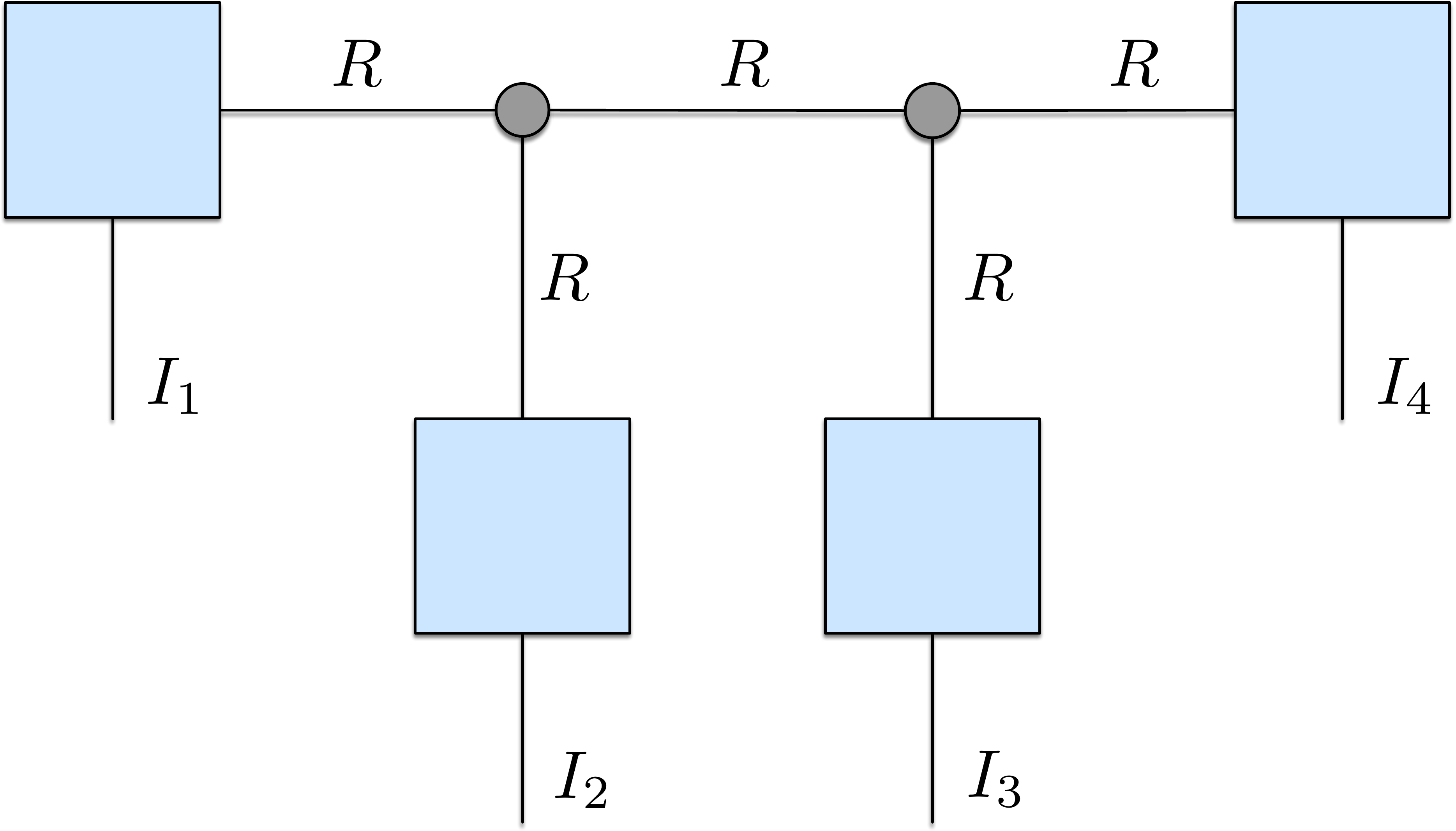}
        \caption{4D CP}
        \label{fig:4dcp}
    \end{subfigure}
    \hfill
	\begin{subfigure}[b]{0.4\textwidth}
        \includegraphics[width=1\linewidth]{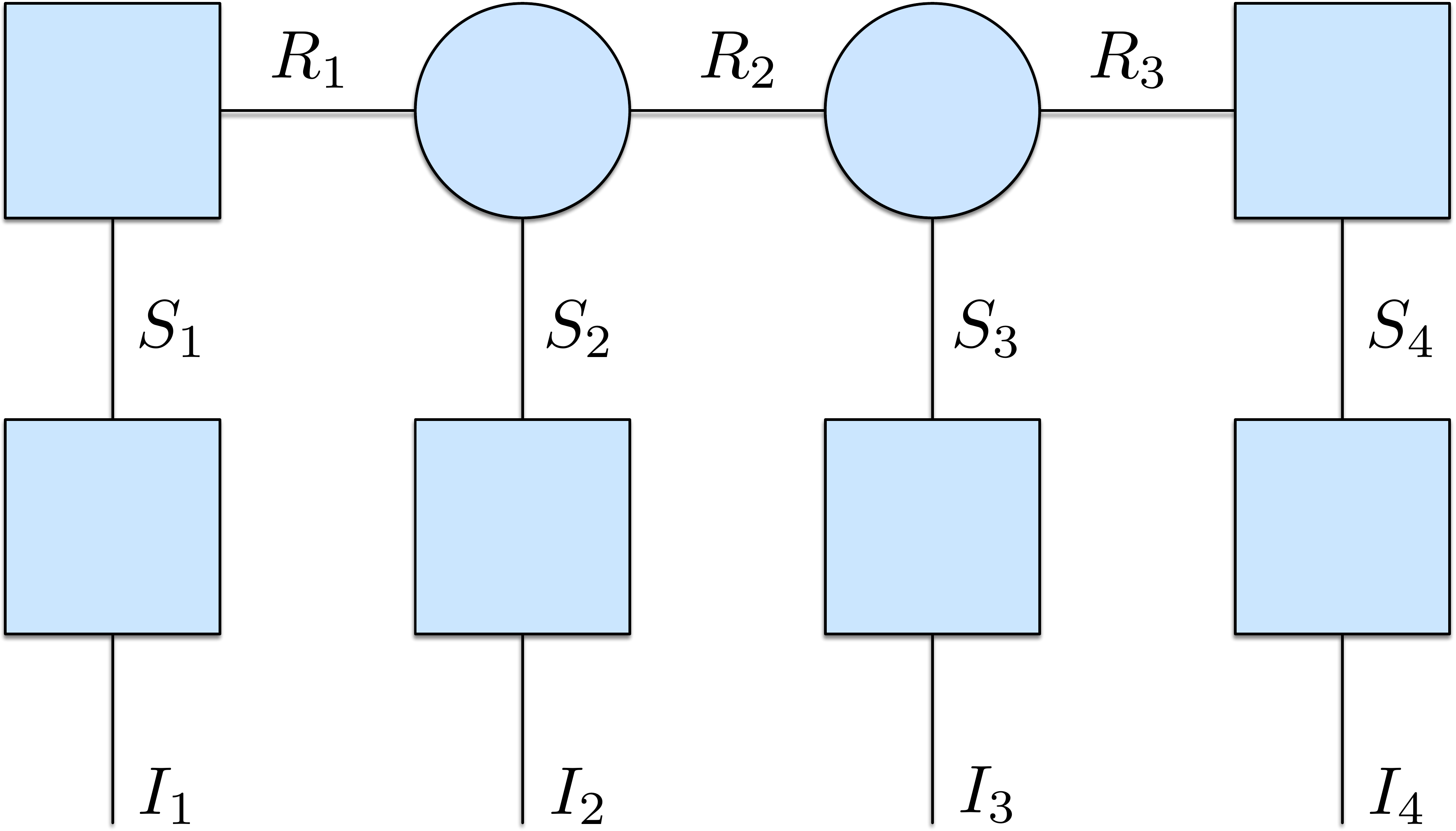}
        \caption{4D TT-Tucker}
        \label{fig:4dtttucker}
    \end{subfigure}
	\begin{subfigure}[b]{0.5\textwidth}
	    \vspace{1.5em}
        \includegraphics[width=1\linewidth]{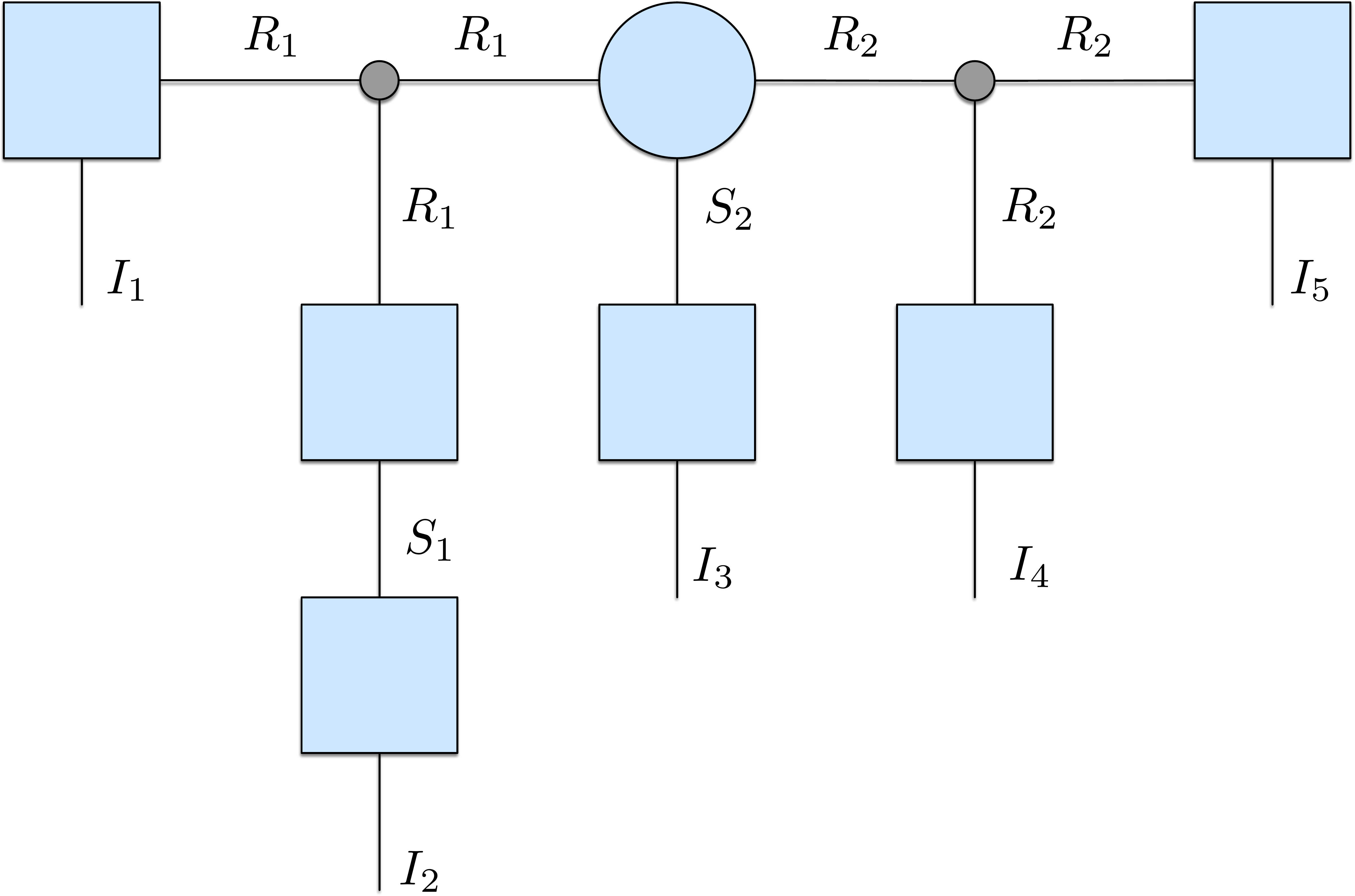}
        \caption{5D CP-TT-Tucker (one of many possible combinations)}
    \end{subfigure}
    \vspace{0.5em}
	\caption{Examples of tensor networks that can be assembled in \acrotntorch{}. A square represents a matrix, as used for CP factors, Tucker factors, and for the first and last cores of a TT; a circle is a 3D tensor and is used for internal TT cores; a dot is a 3D copy tensor~\citep{biamonte2019lectures}, i.e., a super-diagonal tensor with ones along the diagonal.}
	\label{fig:formats}
\end{figure}

\acrotntorch{} supports several decomposition models that are important in the context of machine learning, including \textsc{Candedomp/Parafac}~\citep[CP,][]{Harshman:70}, the Tucker decomposition\footnotemark{}~\citep{LMV:00b}, and the tensor train~\citep[TT,][]{Oseledets:11}. These formats are encapsulated into a single class \texttt{Tensor} and share a common interface for all supported operations, which in turn replicate the API of PyTorch as closely as possible. Internally, the decomposition is stored as a low-rank \emph{tensor network}: a graph whose nodes are low-dimensional tensors and whose edges are tensor dimensions that may be contracted together~\citep{CLOPZM:16} by performing the corresponding operation (e.g., a matrix-vector product). Tensors in \acrotntorch{} can mix more than one format, e.g., one may attach Tucker factors to TT cores, interleave CP and TT cores, or even blend all three formats in various ways. See \cref{fig:formats} for illustrative examples.

\footnotetext[2]{The Tucker tensor is represented as a TT with arbitrary ranks, which has equal expressive power. Furthermore, Tucker factors are implemented as standard unconstrained matrices, in contrast to other implementations that often implement them as unitary/orthogonal matrices.}

In \acrotntorch{}, every decomposition consists of a sequence of cores and, optionally, one or more factors. All nodes of the resulting tensor network may also be accessed directly if the user desires low-level control. % To make operations format-agnostic, we exploit PyTorch's \texttt{einsum} function, a versatile tool for tensor algebra and contraction (merging of nodes along connecting edges), based on Einstein summation algebra.
%In order to make operations format-agnostic, we heavily exploit the versatility of Einstein summation algebra as made possible by PyTorch's \texttt{einsum} function. Effectively, \texttt{einsum} allows to represent tensor operations like summation and multiplication with expressions of tensor elements indices.
By virtue of this abstraction, other tensor formats such as \textsc{Indscal, Candelinc, Dedicom}, and \textsc{Paratuck2}~\citep{KB:09} can be represented as \texttt{Tensor} objects, too. See documentation
for details.

% \begin{itemize}
% 	\item CP: the cores are all 2D (representing CP's factor matrices).
% 	\item TT: the cores are all 3D (representing TT cores).
% 	\item Tucker: the cores are TT-like, but they are chosen so as to be error-free (i.e. they are not rank reduced). In addition, there is a sequence of 2D factors.
% \end{itemize}

% This allows to \emph{blend} these formats in many combinations, for example:

% \begin{itemize}
% 	\item CP-Tucker (\FIXME{reference}): 2D cores are accompanied by factor matrices.
% 	\item TT-Tucker (\FIXME{reference}): 3D cores are accompanied by factor matrices.
% \end{itemize}

\section{Features and Operations}

\subsection{Learning Tensors}
Given a data tensor, \acrotntorch{} can fit a decomposition in multiple ways. E.g., gradient descent is available out-of-the-box, thanks to PyTorch's automatic differentiation and the numerous optimizers it provides. It is very general, as it can be used to learn incomplete tensors, tensors with constraints, or to add various loss terms. For the case of learning TT tensors, a cross-approximation routine is provided that performs adaptive sampling~\citep{OT:10}. Other methods include \emph{TT-SVD}~\citep{Oseledets:11}, \emph{higher-order SVD} and \emph{alternating least squares}~\citep{KB:09} for the TT, Tucker and CP formats, respectively.

\subsection{Tensor Arithmetics}
The library supports tensor$\times$matrix and tensor$\times$vector products, element-wise operations, dot products, convolution, concatenation, mode reordering, padding, orthogonalization, rank truncation, and more. These are useful for common use cases in machine learning: tensor completion; learning low-rank layers or compressing/approximating existing layers; applying multi-linear operators; dimensionality reduction; etc. Advanced element-wise functions such as \texttt{/}, \texttt{exp}, and \texttt{tanh} can be approximately computed via cross-approximation (currently available for TT tensors). %, which estimates arbitrary black-box functions of tensors using an adaptive sampling heuristic and rank selection criterion.
Besides these features, \acrotntorch{} also implements further, miscellaneous tools, such as: sensitivity analysis; Boolean algebra operations; statistical moments; a smart reduction operator; sampling from TT-compressed distributions; multi-linear polynomial expansions; etc.

\subsection{Slicing and Indexing}
Any \texttt{Tensor} may be accessed in several ways:

\begin{itemize}[nosep]
    \item[-] \emph{Basic indexing} via slices in the format \texttt{start:stop:step} and variants, as in \texttt{t[::-1]};
    \item[-] \emph{Fancy indexing} via handcrafted, possibly irregular slices, e.g. by passing a list of indices as in \texttt{t[:, (0, 3, 4)]};
    \item[-] Indexing by NumPy arrays. For example, accessing an $N$-dimensional tensor by an $M \times N$ NumPy matrix returns a 1D tensor with $M$ elements.
    \item[-] Insertion of dummy dimensions, as in \texttt{t[None, :]}, or ellipsis, as in \texttt{t[..., 3]};
    \item[-] Broadcasting is supported, i.e., operations will be repeated along one or more dimensions when their sizes do not match. For example, if \texttt{t} is a 1D tensor, then \texttt{t[:, None] + t[None, :]} is a square matrix.
\end{itemize}
Such read operations match the behavior of PyTorch tensors almost exactly\footnotemark{}. They are performed in the compressed domain and return a compressed \texttt{Tensor}. Moreover, tensors can be edited using the expected syntax, as in \texttt{a[:, 0, :]\ = 2 * b[0, :, :]} or \texttt{t[(0, 2), ..., -1, ::3]\ +=\ 1}. All read/write operations preserve differentiability, a behavior at present unique to \acrotntorch{}, to the best of our knowledge.

\footnotetext[3]{Interleaving fancy and basic indices is not supported, as it is in general computationally expensive.}

\subsection{Batched Tensors}
Batch processing is an important capability to increase the efficiency of tensor-based learning. Besides being differentiable and GPU-ready, decomposition and arithmetic operations in \acrotntorch{} work on batches of tensors in any format; internally, a batch is represented as an extra dimension in every node of the tensor network. As an example, we can decompose multiple tensors into TT cores at once using the TT-SVD algorithm~\citep{Oseledets:11}, with linear algebra operations applied in batches. See \cref{sec:performance} for a benchmark.\\ %Batched tensors are objects of the same class \texttt{Tensor}, 

\subsection{Tensor Train Matrices}
The TT matrix decomposition has recently become a particularly relevant format, because it is, among others, well-suited to compress neural weight layers~\citep{NPOV:15}. Unlike standard tensors, a TT matrix has a different indexing scheme, with two sets of row and column indices. As a consequence, TT matrices come with a distinct set of computation rules for standard operations such as matrix$\times$vector multiplication. Therefore, they have their own class \texttt{TTMatrix}, with a separate API. See additional details in \cref{app:tt_matrix}

\section{Performance}
\label{sec:performance}

We benchmark \acrotntorch{}'s running times across four representative operations: TT decomposition using the TT-SVD algorithm, cross-approximation, and two arithmetic operations that can be achieved by direct manipulation of TT cores~\citep{Oseledets:11}.
%Tensors compressed in the formats supported by \acro{} allow for four core operations without costly decompression.
%
We test four modalities: CPU vs.\ GPU, and in both cases \textit{for loop} vs.\ vectorized batch processing.
As a baseline, we also compare with the Python library ttpy~\citep{ttpy}, which is written in NumPy and \textsc{FORTRAN} and also implements these four operations.
All experiments use randomly initialized tensors of TT-rank $R = 20$, physical dimension sizes $I = 15, \dots, 45$, and number of dimensions $N = 8$ (except for the TT-SVD experiment, where $N = 4$).
We used PyTorch 1.13.0a0+git87148f2 (compiled from source) and NumPy 1.22.4 on an Intel(R) Core(TM) i7-7700K CPU with 64Gb RAM and an NVIDIA GeForce RTX 3090 GPU.

\begin{figure}[!t]
	\centering
	\begin{subfigure}[b]{0.49\textwidth}
        \includegraphics[width=1\linewidth]{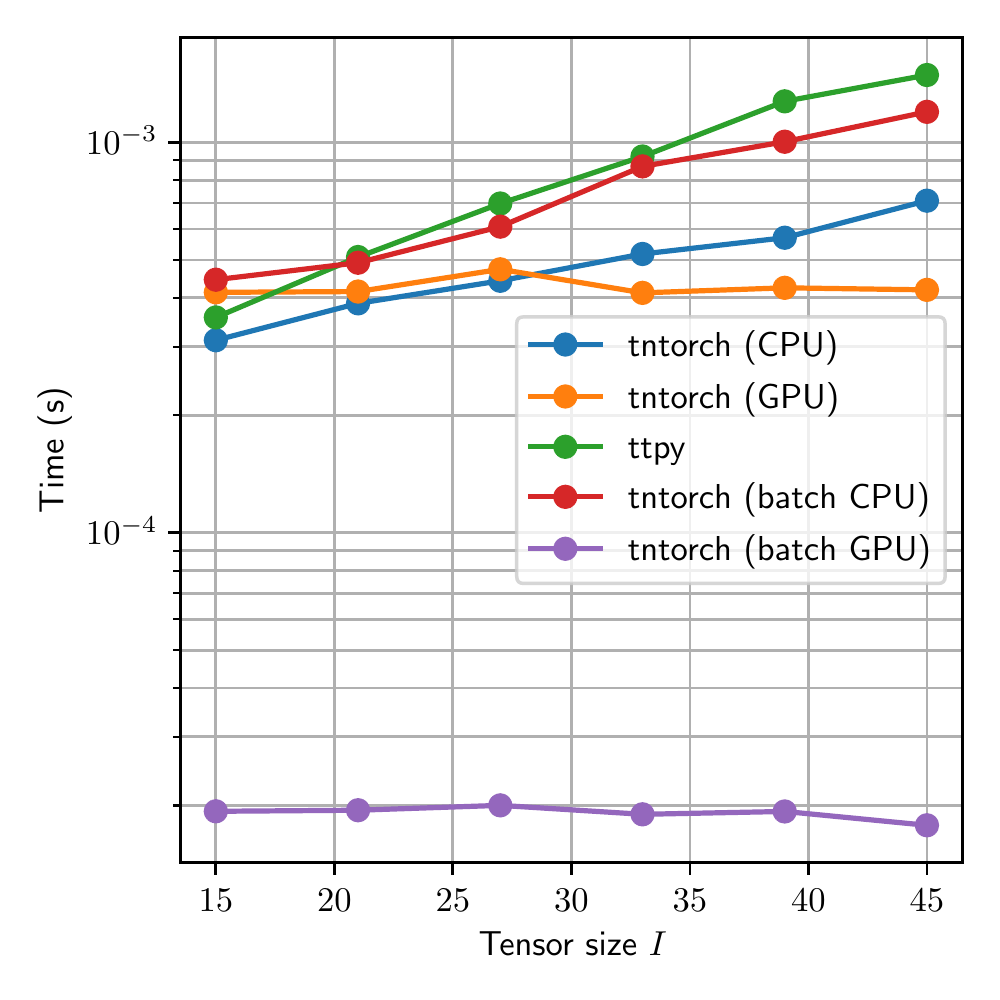}
        \caption{Element-wise sum}
    \end{subfigure}
    \hfill
	\begin{subfigure}[b]{0.49\textwidth}
        \includegraphics[width=1\linewidth]{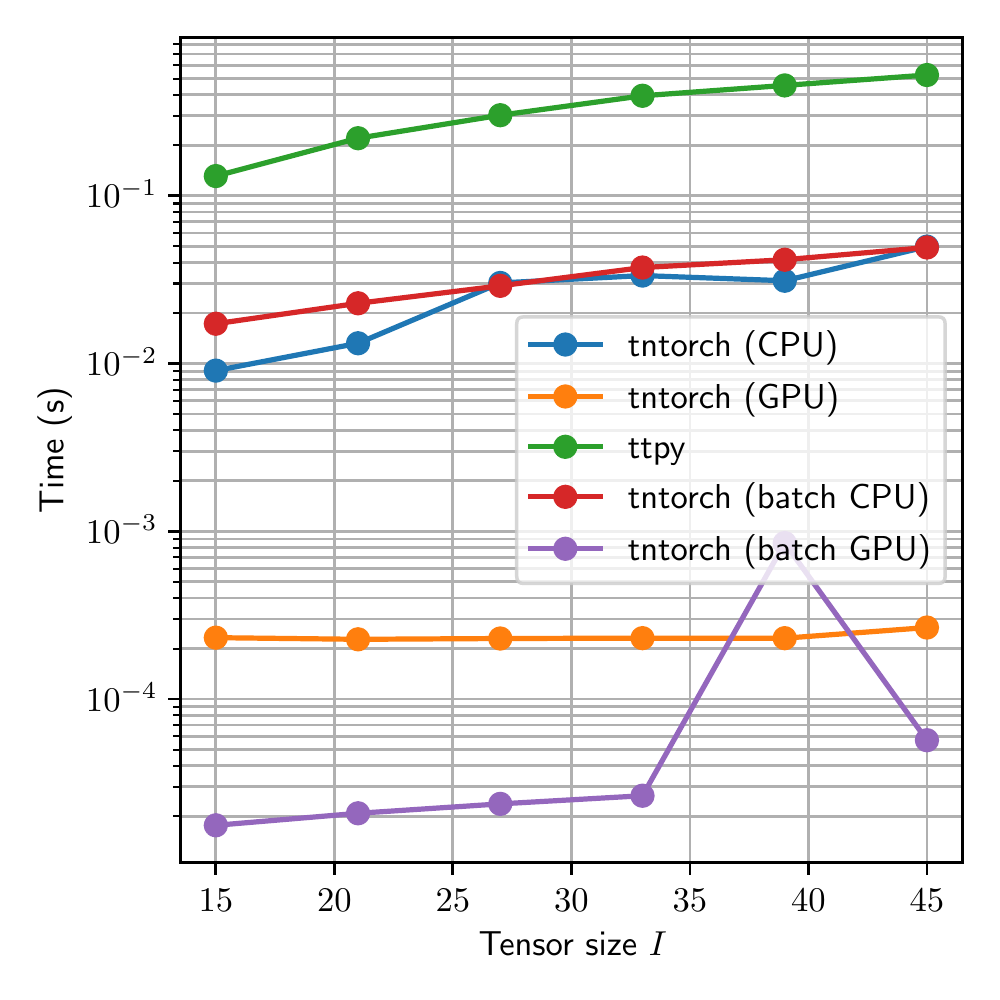}
        \caption{Element-wise product}
    \end{subfigure}
	\begin{subfigure}[b]{0.49\textwidth}
    \vspace{1em}
        \includegraphics[width=1\linewidth]{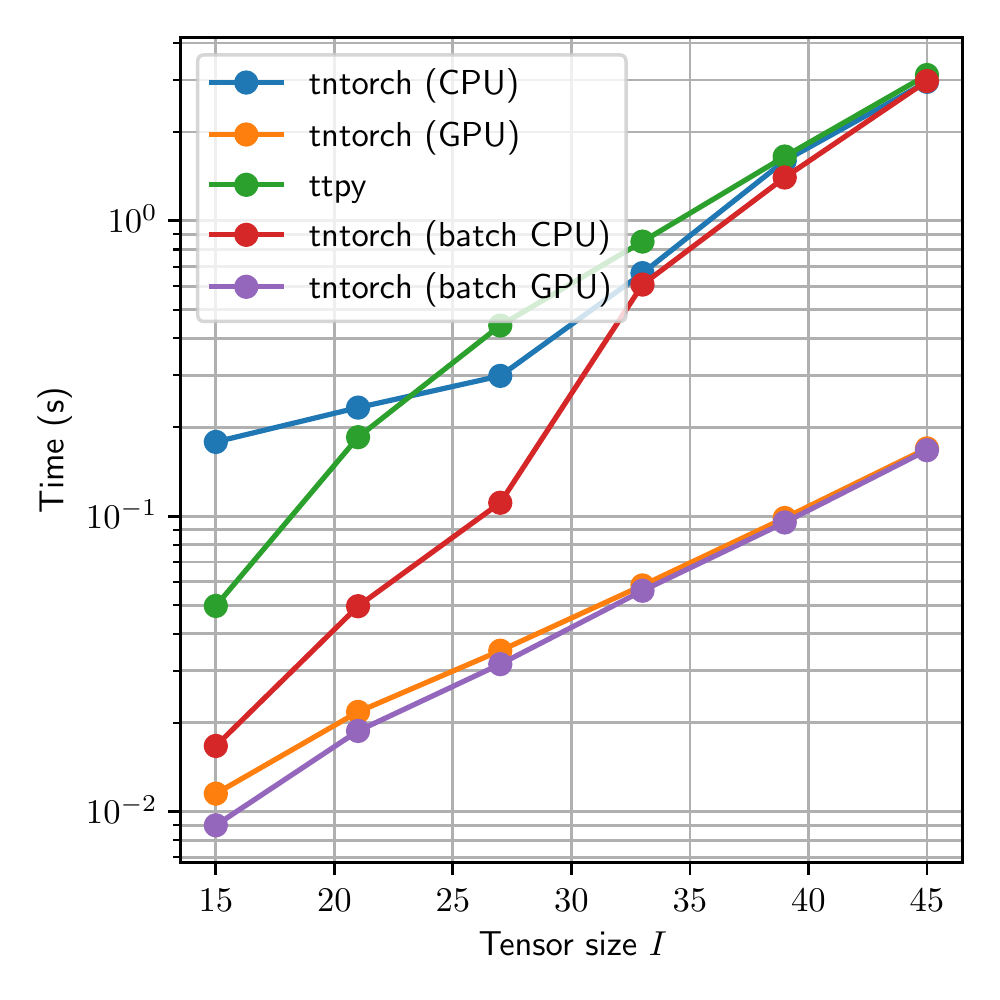}
        \caption{TT-SVD}
    \end{subfigure}
    \hfill
	\begin{subfigure}[b]{0.49\textwidth}
	    \vspace{1em}
        \includegraphics[width=1\linewidth]{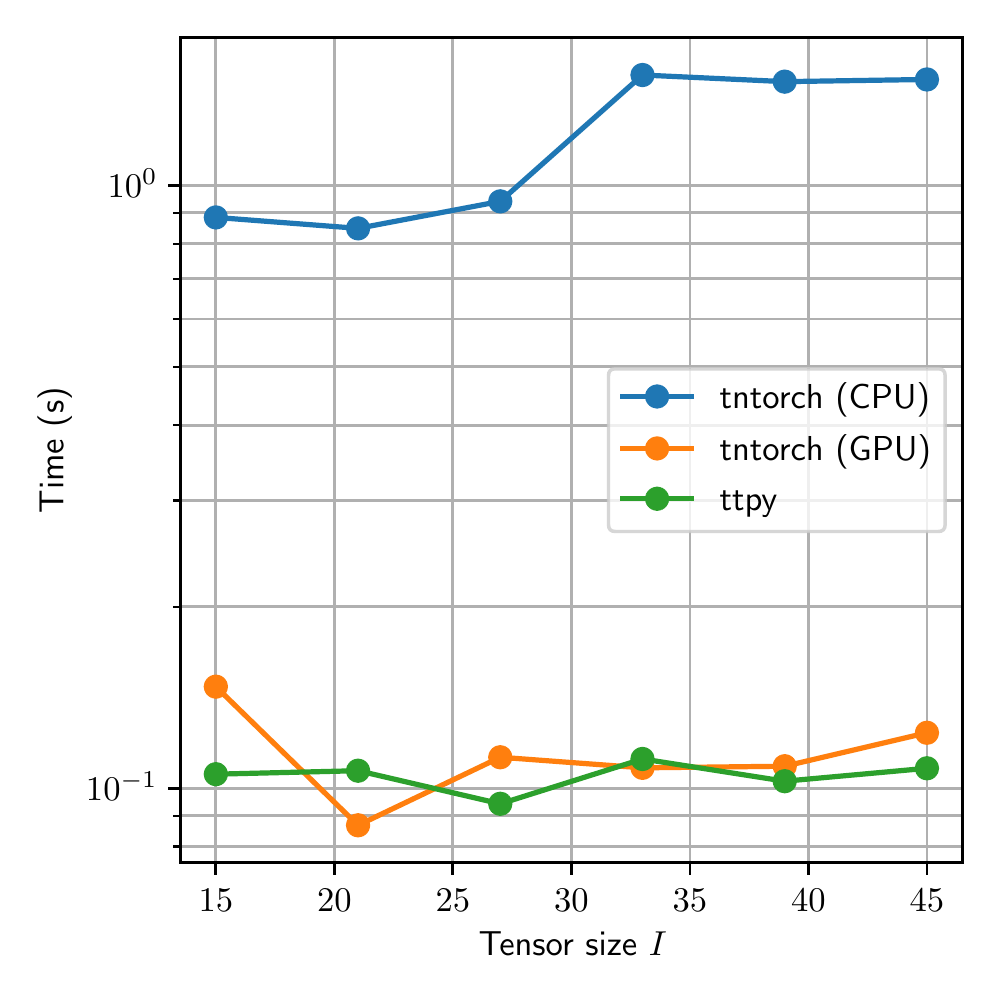}
        \caption{Cross-approximation}
    \end{subfigure}
        \vspace{0.5em}
	\caption{Runtimes of four operations with varying tensor sizes (average time per processed object, over 10 runs). The batch size is set to $B = 32$. Best viewed in color.}
	\label{fig:performance}
\end{figure}

Results are reported in \cref{fig:performance}. Note that the GPU is more performant on both batch and non-batch modes. Also, \acrotntorch{} scales better (or similarly, for cross-approximation) than the baseline w.r.t.\ the tensor size, and is thus a good fit for data-intensive ML applications.
% For element-wise summation and product batch implementation on GPU is superior in runtime compared to all other methods.
% %
% ttpy performs element-wise summation more efficiently for smaller tensor sizes but for larger tensors with tensor size greater than 15 \acro{} implementation becomes more efficient.
% %
% \acro{} GPU implementation only becomes more efficient with tensor size grater than 25.
% %
% For element-wise product \acro{} is faster compared to ttpy, GPU allows for one order of magnitude speed-up.
% %
% For TT-SVD \acro{} outperforms ttpy for tensor sizes greater than 10.
% %
% Cross-approximation demonstrates comparable performance for both ttpy and \acro{}.

\section{Conclusions}

We have introduced \acrotntorch{}, a PyTorch-powered library that unifies multiple efficient tensor formats under the same interface, together with a rich suite of learning and analysis routines. The library comes with many standard features of modern machine learning frameworks, including auto-differentiation, GPU and batch processing, and advanced indexing. \acrotntorch{} imitates the look and feel of standard PyTorch tensors, while making the power of low-rank tensor decompositions accessible for machine learning.

\clearpage
\appendix
\section{Appendix}

\subsection{Additional details}
We heavily exploit the versatility of Einstein summation algebra as made possible by PyTorch's \texttt{einsum} function. Effectively, \texttt{einsum} allows to represent tensor operations like summation and multiplication using compact strings of symbols, each encoding one tensor index.
\acrotntorch{} represents basic decompositions using:
\begin{itemize}
	\item CP: a sequence of 2D factor matrices
	\item TT: a sequence of 3D tensor train cores
	\item Tucker: the cores are TT-like, but they are chosen so as to be error-free (i.e. they are not rank reduced). In addition, there is a sequence of 2D factors to encode the Tucker bases.
\end{itemize}

\subsection{Comparison between tensor libraries}
\label{app:comparison}
\begin{table*}[h!]
    \centering
    \footnotesize
    \setlength\tabcolsep{4.4pt}
	\begin{tabular}{ cccccccc } 
		%\hline
		& \acrotntorch{}  & T3F & TensorLy & TedNet & TensorD & TensorNetwork & tt-pytorch \\
		\hline
		Tensor arithmetics & \checkmark & \checkmark & - & - & \checkmark & - & - \\
		\rowcolor{gray!12}
		Fancy indexing & \checkmark & - & - & - & - & - & - \\
		CP decomposition & \checkmark & - & \checkmark & \checkmark & \checkmark & \checkmark \footnotemark [1] & - \\
		\rowcolor{gray!12}
		Tucker decomposition & \checkmark & - & \checkmark & - & \checkmark & \checkmark \footnotemark [1] & -  \\
		TT decomposition & \checkmark & \checkmark & \checkmark & \checkmark & - & \checkmark \footnotemark [1] & \checkmark \\
		\rowcolor{gray!12}
		CP/Tucker/TT blending & \checkmark \footnotemark [2] & - & - & - & - & \checkmark \footnotemark [1] & - \\
		Tensor ring & - & - & - & \checkmark & - & \checkmark \footnotemark [1] & \checkmark \\
		\rowcolor{gray!12}
		Hierarchical/tree formats & - & - & - & - & - & \checkmark \footnotemark [1] & - \\
		TT matrix & \checkmark & \checkmark & \checkmark & - & - & \checkmark \footnotemark [1] & - \\
		\rowcolor{gray!12}
		CP matrix & \checkmark & - & - & - & - & \checkmark \footnotemark [1] & - \\
		Cross-approximation & \checkmark & - & \checkmark & - & - & - & - \\
		\rowcolor{gray!12}
		Global optimization & \checkmark & - & - & - & - & - & - \\
		Riemannian optimization & - & \checkmark & - & - & - & - & - \\
		\rowcolor{gray!12}
		Batch processing & \checkmark & \checkmark \footnotemark [2] & - & \checkmark & - & \checkmark \footnotemark [1] & - \\
		Sparse tensor algorithms & - & \checkmark & \checkmark & - & - & - & - \\
		\hline
	\end{tabular}
	\caption{Feature comparison between \acrotntorch{} and six related libraries that also support GPU processing and automatic differentiation as of June 2022.}
	\label{tab:comparison}
\end{table*}
\footnotetext[1]{Tensor networks can be constructed, but no tools are provided to decompose a given data tensor into a desired network.}
\footnotetext[2]{Batched decomposition of a data tensor is not supported.}

\subsection{Cross-approximation}
\label{subsec:cross}

% \begin{algorithm}[h!]
% 	\centering
% 	\caption{\textsc{Cross-approximation}}
% 	\begin{algorithmic}[1]
% 		\vspace*{.18cm}
% 		\STATEx \hspace*{-.6cm} $\sI = I_1 \times \cdots \times I_d$ -- $d$-dimensional grid 
% 		\STATEx \hspace*{-.6cm} $\tX \in \R^{\sI}$ -- input data
% 		\vspace*{.1cm}
% 		\REQUIRE Input data $\tX$
% 		\FOR{$\text{epoch} = 1, \dots, n_{\text{iterations}}$}
% 		\FOR{$k = 1, \cdots, d$}
% 		\STATE Select CA indices for $k$-th fiber using \emph{maxvol} $J^k_{CA} \subset \sI$
% 		\ENDFOR
% 		\ENDFOR
% 		\STATEx {\bfseries Output:} $
% 		\tQ_1, \displaystyle
% 		\cdots,
% 		\tQ_d$
% 	\end{algorithmic}
% 	\label{alg:ca}
% \end{algorithm}

\acrotntorch{} implements \emph{cross-approximation} \citep{OT:10}, a powerful technique to learn a compressed TT tensor out of a black-box function by evaluating it on a limited set of samples. This is a valuable tool when the target tensor does not fit into memory but it can be sampled on demand, and can be used for surrogate modeling or to efficiently compute functions of already compressed tensors. Cross-approximation has also been adapted as a gradient-free global optimizer for discrete problems, and an implementation is available in \acrotntorch{}. See \cite{sozykin2022ttopt} for a detailed example of an application in reinforcement learning. \acrotntorch{} also provides a differentiable version of cross-approximation which alternates between sample selection (which is a combinatorial, non-differentiable heuristic), followed by evaluation and backpropagation of the target function on the selected samples \citep{UMBRKS:21}.

Under the hood, cross-approximation selects indices based on a tensor version of an earlier algorithm known as \emph{maxvol} \citep{MO:15}. In short, \emph{maxvol} is a greedy heuristic that finds the $k$ most linearly independent rows or columns from a given matrix (known to be an $\mathcal{NP}$-hard problem \citep{civril2009selecting}). We use the \emph{maxvolpy} library by \cite{maxvolpy}, which relies on BLAS, for this algorithm, but \acrotntorch{} also provides its own implementation to fall back to whenever BLAS is not available.

\subsection{Tensor Train Matrices}
\label{app:tt_matrix}
\begin{figure}[h!]
	\centering
	\begin{subfigure}[b]{0.49\textwidth}
	    \centering
        \includegraphics[scale=0.2]{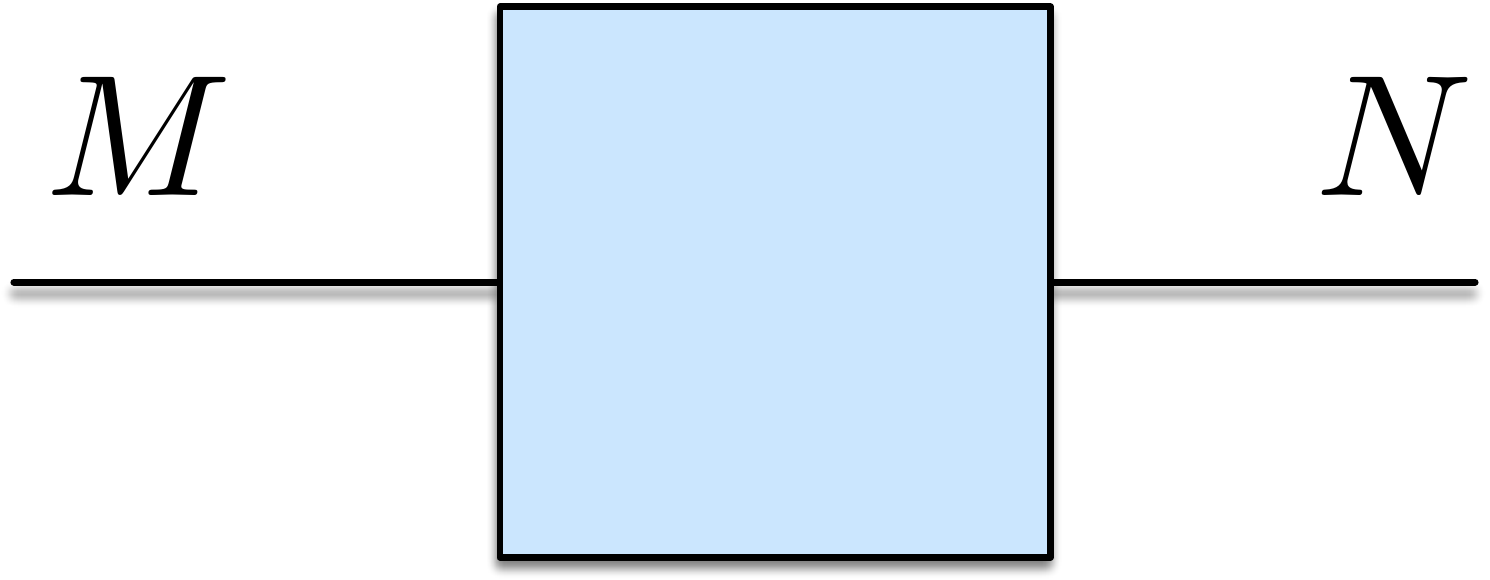}
        \caption{Matrix of shape $M\times N$}
        \label{fig:matrix}
    \end{subfigure}
    \hfill
    \begin{subfigure}[b]{0.49\textwidth}
	    \centering
        \includegraphics[scale=0.2]{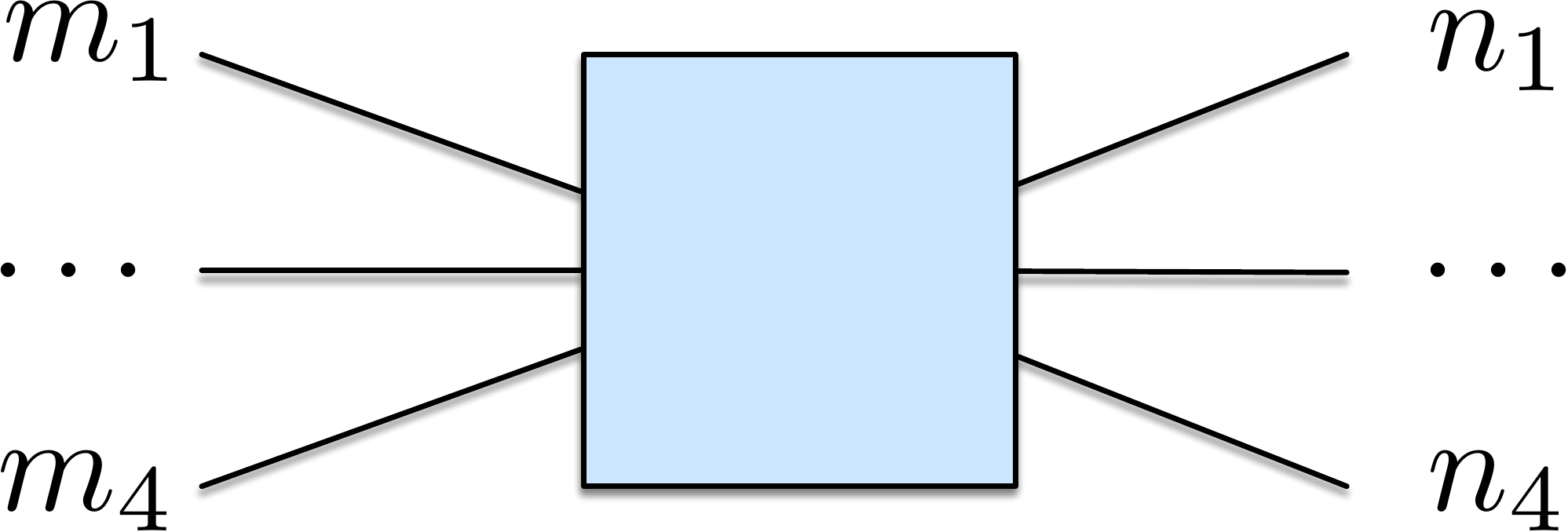}
        \caption{Tensor of shape $m_1 \times n_1 \times \dots \times m_4 \times n_4$}
        \label{fig:resh1}
    \end{subfigure}
	\begin{subfigure}[b]{0.49\textwidth}
	    \centering
        \includegraphics[scale=0.2]{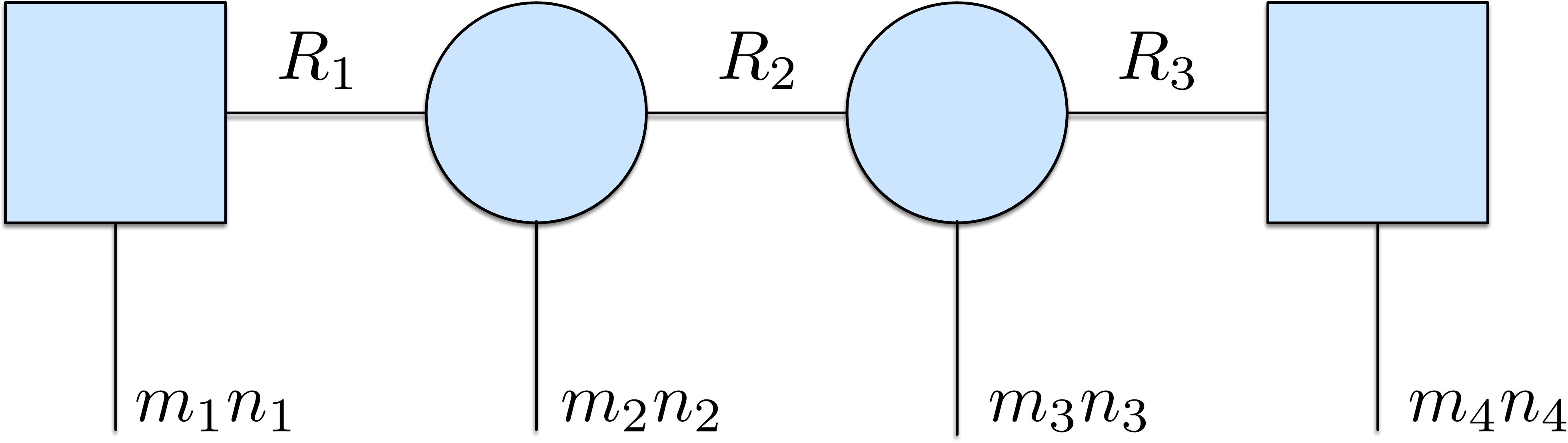}
        \caption{4D TT tensor}
        \label{fig:ttm1}
    \end{subfigure}
    \begin{subfigure}[b]{0.49\textwidth}
        \centering
        \includegraphics[scale=0.2]{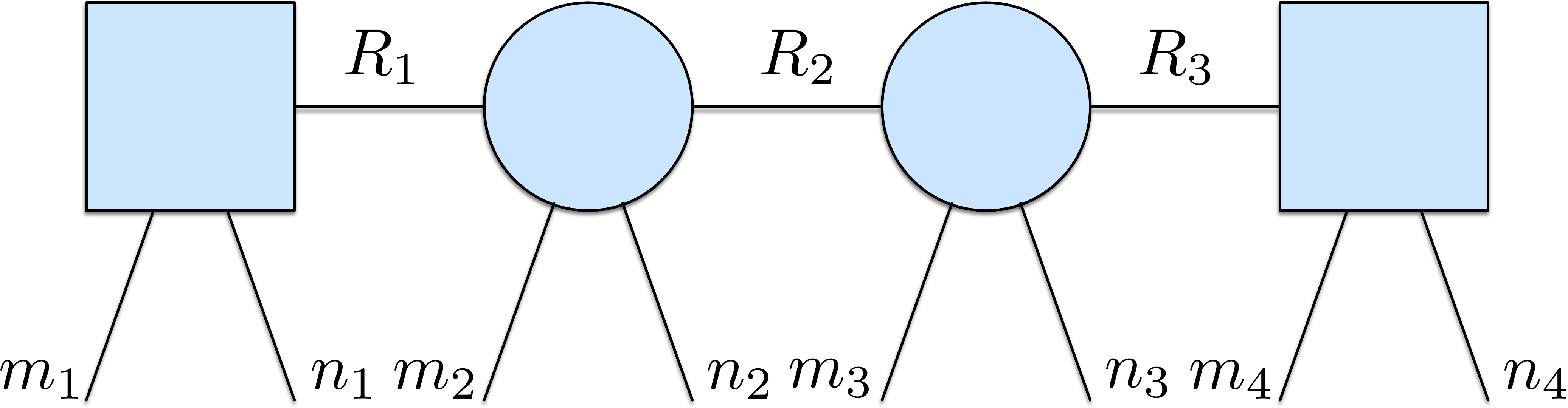}
        \caption{4D TT matrix}
        \label{fig:ttm2}
    \end{subfigure}
	\caption{Example of conversion of matrix to 4D TT tensor and corresponding TT matrix}
	\label{fig:ttm_tt}
\end{figure}

A matrix of shape $M\times N$ (\cref{fig:matrix}) transforms to a tensor of shape $m_1 \times \dots \times m_k \times n_1 \times \dots \times n_k$, where $M = m_1 \times \dots \times m_k$ and $N = n_1 \times \dots \times n_k$. Corresponding row and column dimensions are grouped together and the shape of the resulting tensor becomes $m_1 \times n_1 \times \dots \times m_k \times n_k$ (\cref{fig:resh1}). The tensor is subject to standard TT decomposition with the cores of shape $R_i \times m_i n_i \times R_{i + 1}$ (\cref{fig:ttm1}) and finally each core is reshaped to $R_i \times m_i \times n_i \times R_{i + 1}$ (\cref{fig:ttm2}), separating each input and output dimension on the core level.
Many operations discussed in \cite{izmailov2018scalable} have also been implemented in \acrotntorch{}, including fast matrix inverse and determinant algorithms for rank-1 TT matrices that are equivalent to Kronecker products.

\begin{figure}[h!]
	\centering
	\begin{subfigure}[b]{0.49\textwidth}
	    \centering
        \includegraphics[scale=0.2]{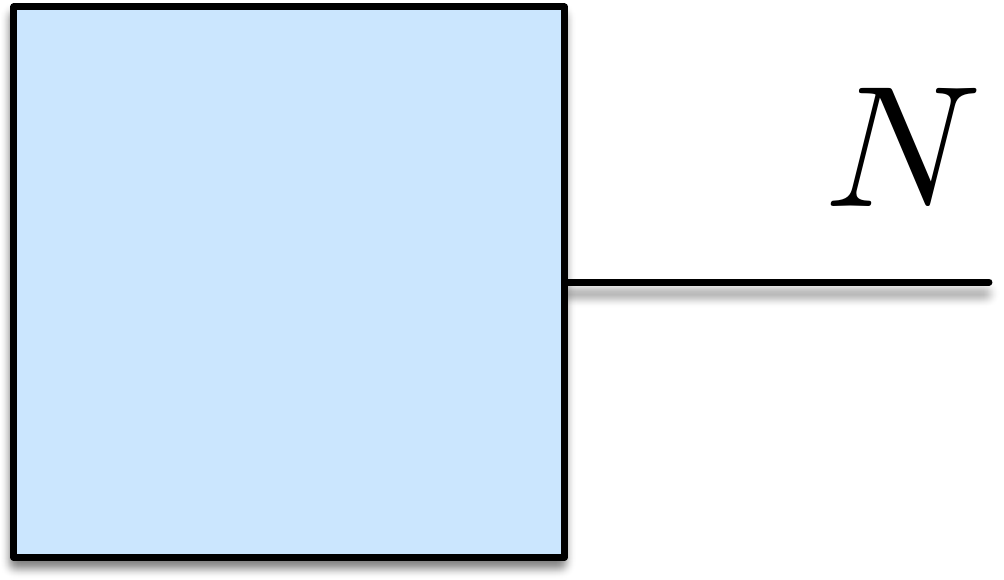}
        \caption{Vector of shape $N$}
        \label{fig:vector}
    \end{subfigure}
    \hfill
    \begin{subfigure}[b]{0.49\textwidth}
	    \centering
        \includegraphics[scale=0.2]{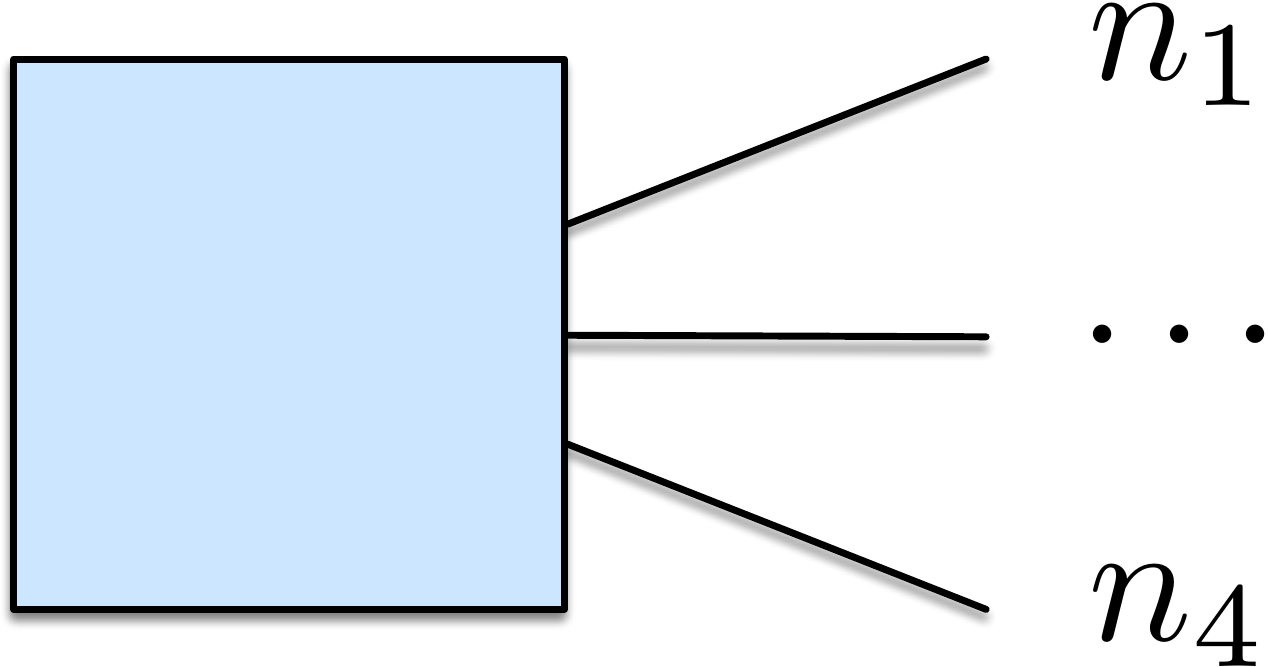}
        \caption{Vector of shape $n_1\times \dots \times n_4$}
        \label{fig:resh2}
    \end{subfigure}
	\caption{Example of vector transformation for \texttt{tt\_multiply}}
	\label{fig:ttm_multiply}
\end{figure}

\texttt{tt\_multiply}.~~In TT format matrix$\times$vector multiplication does not require decompression. The vector of shape $N$ (\cref{fig:vector}) is reshaped as in \cref{fig:ttm_multiply} to a tensor of shape $n_1\times \dots \times n_k$ (\cref{fig:resh2}), such that $N = n_1\times \dots \times n_k$ and corresponding dimensions are contracted with the TT matrix (\cref{fig:ttm2}). The result of matrix$\times$vector operation is a contracted tensor of shape $m_1\times \dots \times m_k$, which can be reshaped back to a vector of shape $M$. See \citep{NPOV:15} for more details.

\subsection{CP Matrices}
\label{app:cp_matrix}
\begin{figure}[h!]
	\centering
    \includegraphics[height=100pt]{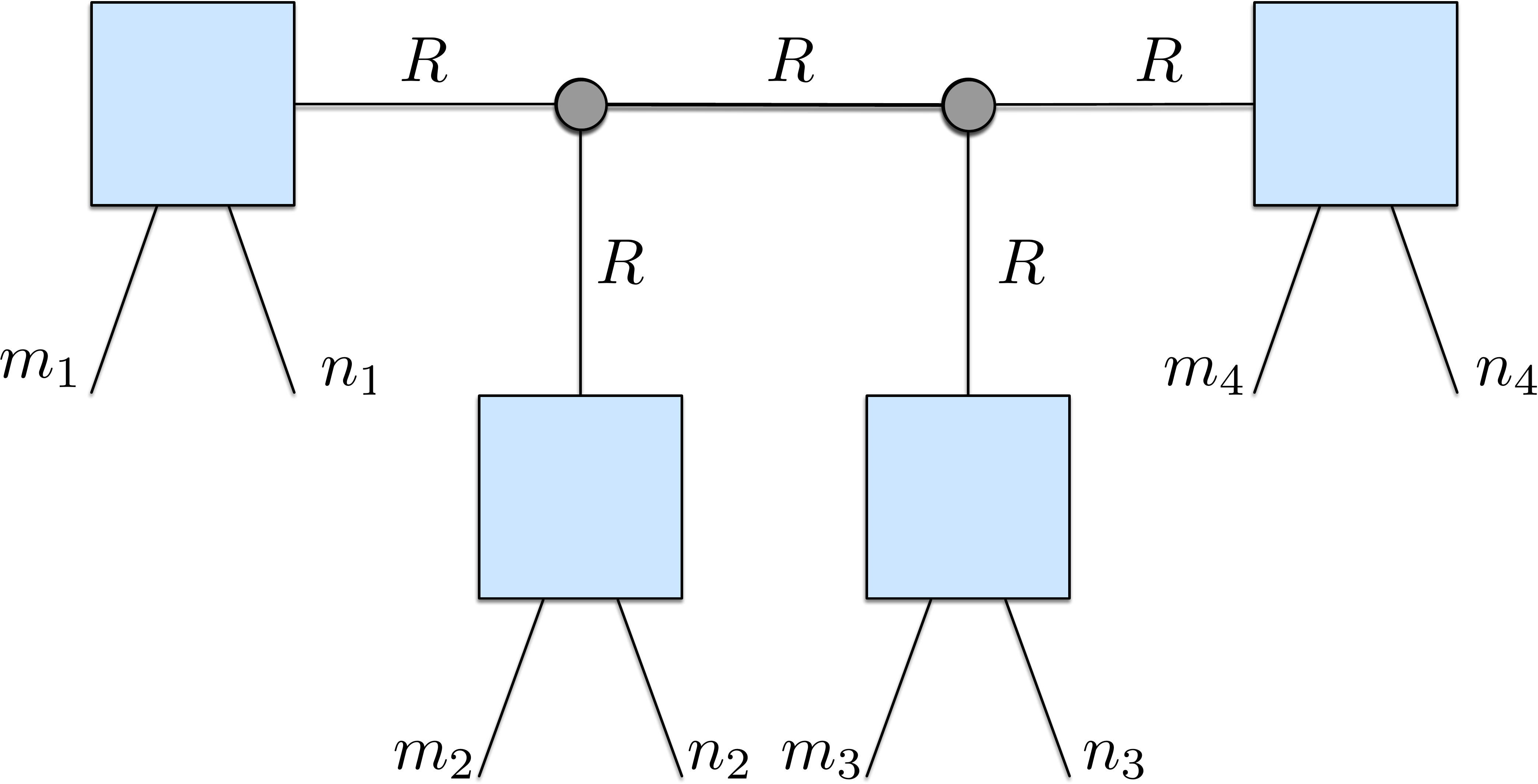}
    \caption{4D CP matrix, during matrix$\times$vector multiplication dimensions $m_k$ are contracted with corresponding vector dimensions}
	\label{fig:cpm}
\end{figure}

Similarly to TT matrix decomposition, we have also implemented a CP matrix decomposition. The scheme for indexing and regrouping dimensions is identical to that of \texttt{TTMatrix}. In a CP matrix, each CP factor has shape $m_i n_i \times R$ but is conceptually equivalent to $m_i \times n_i \times R$, where $m_i$ and $n_i$ are spatial dimensions. Like \texttt{TTMatrix}, CP matrices require custom operations (such as matrix$\times$vector products) that are not available for CP tensors and thus are implemented as a separate class \texttt{CPMatrix}. See \cref{fig:4dcp} for a visualization of a CP tensor and \cref{fig:cpm} for the corresponding CP matrix with $I_k = m_k \times n_k$.

\clearpage
\bibliographystyle{apalike}
\bibliography{references}

\end{document}